\documentclass[10pt]{article}

\usepackage[letterpaper,margin=1in]{geometry}
\usepackage[T1]{fontenc}
\usepackage[utf8]{inputenc}
\usepackage{lmodern}
\usepackage{microtype}
\usepackage{amsmath,amssymb,amsthm,mathtools}
\usepackage{bm}
\usepackage{graphicx}
\usepackage{subcaption}
\usepackage{booktabs}
\usepackage{multirow}
\usepackage{array}
\usepackage{enumitem}
\usepackage{xcolor}
\usepackage{siunitx}
\usepackage{url}
\usepackage[numbers,sort&compress]{natbib}
\usepackage[colorlinks=true,linkcolor=blue,citecolor=blue,urlcolor=blue]{hyperref}
\usepackage[nameinlink,capitalise,noabbrev]{cleveref}
\usepackage{titlesec}
\usepackage{fancyhdr}
\usepackage{abstract}
\usepackage{fontawesome5}

\setlist[itemize]{topsep=2pt,itemsep=1pt,leftmargin=1.5em}
\setlist[enumerate]{topsep=2pt,itemsep=1pt,leftmargin=1.7em}
\titleformat{\section}{\normalsize\bfseries\scshape}{\thesection}{0.55em}{}
\titleformat{\subsection}{\normalsize\bfseries}{\thesubsection}{0.5em}{}
\titleformat{\subsubsection}{\normalsize\itshape}{\thesubsubsection}{0.5em}{}
\titlespacing*{\section}{0pt}{0.85em}{0.3em}
\titlespacing*{\subsection}{0pt}{0.65em}{0.25em}
\titlespacing*{\subsubsection}{0pt}{0.5em}{0.2em}

\hypersetup{
  pdftitle={Designing Agent-Ready Websites for AI Web Agents: A Framework for Machine Readability, Actionability, and Decision Reliability},
  pdfauthor={Said Elnaffar and Farzad Rashidi},
  pdfsubject={AI web agents; agent-ready websites; e-commerce},
  pdfkeywords={AI web agents, AI-enhanced e-commerce, smart commerce systems}
}

\theoremstyle{definition}

\theoremstyle{remark}

\renewcommand{\thefootnote}{\ifnum\value{footnote}=2 \ensuremath{\dagger}\else\arabic{footnote}\fi}

\newcommand{\email}[1]{\href{mailto:#1}{\nolinkurl{#1}}}
\newcommand{\papertitle}{Designing Agent-Ready Websites for AI Web Agents: A Framework for Machine Readability, Actionability, and Decision Reliability\footnote{Replication Packages: \href{https://github.com/rashidiff/AIAgentReadWebSites}{\faGithub}}}

\newcommand{\makecustomtitle}{%
  \begin{center}
    {\large\bfseries \papertitle\par}
    \vspace{0.8em}
    \noindent\rule{\textwidth}{0.5pt}\par
    \vspace{0.9em}
    \begin{minipage}[t]{0.46\textwidth}
      \centering
      {\normalsize\bfseries Said Elnaffar{\normalfont\,{\scriptsize PhD}}\footnotemark\par}
      \vspace{0.2em}
      {\small Independent Researcher\\
      London, ON, Canada\\
      \email{said.elnaffar@gmail.com}\par}
    \end{minipage}\hfill
    \begin{minipage}[t]{0.46\textwidth}
      \centering
      {\normalsize\bfseries Farzad Rashidi\par}
      \vspace{0.2em}
      {\small UFR Informatique\\
      Université Paris Cité\\
      F-75013 Paris, France\\
      \email{farzad.rashidi@etu.u-paris.fr}\par}
    \end{minipage}\par
    \vspace{1.2em}
    {\small July 13, 2026\par}
  \end{center}
  \footnotetext{Corresponding author.}
}

\begin{document}

\makecustomtitle
\thispagestyle{fancy}

\begin{abstract}
Online shopping is increasingly shifting toward a model in which AI agents independently search for products, compare options, evaluate constraints, and carry out parts of the purchasing process for users. Website design must now support both human and agent-mediated interaction. This paper introduces the agent-ready website, a design framework for enhancing the readability, interpretability, verifiability, and actionability of e-commerce platforms for AI agents. Existing web design, SEO, and generative engine optimization (GEO) metrics do not fully assess a website's capacity for agent-mediated interaction. The proposed framework is structured around three dimensions agent interpretability, agent executability, and agent decision reliability supported by features such as machine readability, semantic clarity, agent actionability, and contextual decision-reliability signals. The framework is evaluated through a controlled experiment comparing a human-oriented baseline and an agent-ready version of an identical website prototype, with identical catalogs, pricing, stock, and shopping workflows. The evaluation involved five tasks, three browser-agent models (GPT-4.1, Gemini-2.5 Flash, and Grok-4 Fast), and 300 runs, measuring PASS/PARTIAL/FAIL outcomes, strict and functional success rates, error patterns, step counts, and token consumption. The agent-ready website achieved 134 PASS runs out of 150 versus 74 out of 150 for the baseline (strict success rates of 89.3\% vs. 49.3\%), with the largest gains in product detail extraction, comparison, and multi-constraint selection. It also reduced PARTIAL outcomes from 43 to 3 and lowered the average step count from 9.31 to 6.49. These results provide preliminary evidence that enhanced structural clarity, action cues, evidence signals, and temporal validity indicators can substantially improve the reliability and efficiency of AI browser agents.
\end{abstract}

\noindent\textbf{CCS CONCEPTS} \textbullet{} Intelligent agents \textbullet{} Web-based interaction \textbullet{} Electronic commerce

\noindent\textbf{Keywords:} AI web agents, AI-enhanced e-commerce, smart commerce systems
\section{Introduction}
With the rise of AI agents, user interaction with websites is shifting from direct human–website engagement to AI agent–website interaction. In this paradigm, AI agents handle tasks such as searching, comparing, decision-making, and executing actions on websites, effectively replacing human users. To date, website developers have primarily designed for human users, emphasizing graphical user interfaces, page clarity, and SEO-optimized, fast-loading sites. Consequently, professionals in web business and commerce have prioritized SEO rankings, visibility, and site traffic. However, discoverability for human visitors does not inherently ensure that a website is readable, interpretable, or actionable for AI agents. Currently, users delegate activities such as searching, comparing, verifying, and executing online actions to AI agents \cite{jeannot2022visual,deng2023mind2web,garett2016literature,he2024webvoyagerbuildingendtoendweb,NISSEN2024108168,palmer2002web}. This shift indicates that traditional web development standards, previously centered on human-friendliness and SEO, are no longer adequate for AI agents. Therefore, developers must modify their websites to accommodate AI agents in the same way they align them with human users \cite{berman2013role,ERDMANN2022650,articlssssse}.\\

Recent studies suggest that the performance of AI agents in web environments depends not only on the capabilities of the underlying AI model but also on the characteristics of the website itself. To accomplish web-based tasks, a website must have an interpretable structure, transparent information, identifiable interactive elements, and well-defined execution pathways. Nonetheless, research indicates that AI agents continue to face significant challenges in understanding page structure, identifying user interface elements, and executing multi-step interactions across diverse websites \cite{deng2023mind2web, he2024webvoyagerbuildingendtoendweb, yao2023webshopscalablerealworldweb, zheng2024gpt4visiongeneralistwebagent, zhou2024webarenarealisticwebenvironment}. The following section reviews the relevant literature on web agent architectures, failure modes, and the limitations of existing web design paradigms that motivate the proposed framework.

\section{Background and Conceptual Foundations}
\subsection{Web Agents and Interaction Paradigms}
Web AI agents operate in live web environments by sequentially executing actions such as clicking, submitting forms, and navigating to accomplish user-defined tasks, thereby eliminating the need for direct human interaction with websites. AI agents in web environments follow a perception-reasoning-action cycle: they capture the current page state, analyze content and interactive elements, then select the next action most likely to advance the task objective \cite{unknown,deng2023mind2web,unknssown,pmlr-v70-shi17a}. Since each action can affect subsequent steps, completing a web task involves multiple decisions. The variability in website design, dynamic content, multi-step forms, and interactive elements make web environments complex for AI. Success requires combining understanding of page content with precise action execution across diverse scenarios \cite{drouin2024workarena,ssssss,hong2024cogagentvisuallanguagemodel,liu2018reinforcementlearningwebinterfaces,liu2025agentbenchevaluatingllmsagents}.

\subsection{Failure Modes of Web Agents}
Web agents do not fail due to a single cause but through compounded failures across four distinct pipeline stages. Observational failures occur when agents encounter incomplete or structurally inconsistent representations of webpages, because websites are primarily engineered for human perception rather than machine processing. Grounding failures occur when the correct semantic intent is incorrectly associated with an inappropriate user interface element due to interface ambiguity and structural inconsistencies among interactive regions. Execution failures originate from unstable interactions, missed clicks, navigation errors, and authentication interruptions, thereby reflecting the discrepancy between rigid agent actions and the event-driven behavior characteristic of live web applications \cite{deng2023mind2web, ssssss, koh2024visualwebarenaevaluatingmultimodalagents, zheng2024gpt4visiongeneralistwebagent, zhou2024webarenarealisticwebenvironment}. Decision failures, in contrast, result in incorrect action selection despite sufficient information being available. This is attributable to the challenges associated with multi-step planning and the limited capacity for recovery over extended task horizons \cite{hong2024cogagentvisuallanguagemodel, liu2025agentbenchevaluatingllmsagents, xie2024osworldbenchmarkingmultimodalagents, yao2023webshopscalablerealworldweb}. 

\subsection{Limitations of Existing Web Design}
Existing web optimization paradigms, which include visibility-driven search enhancements, interpretability-focused structured data, usability-oriented accessibility standards, and interoperability-based application programming interfaces (APIs), are intrinsically fragmented owing to their emphasis on isolated, human-centered objectives \cite{guha2016schema, hou2026mcp, Wang_2024, yang2023setofmarkpromptingunleashesextraordinary, yao2023reactsynergizingreasoningacting}. Consequently, they do not collectively provide the cohesive foundation necessary for the comprehensive execution of AI agent tasks. This demands that websites support multimodal content observation, clear visual structure, reliable action execution, and sustained decision-making across multi-step interactions. This limitation reflects a structural mismatch: the conventions of traditional web design were not developed with programmatic observation or automated action in mind, and existing interfaces often lack the properties needed \cite{dulacarnold2019challengesrealworldreinforcementlearning, koh2024visualwebarenaevaluatingmultimodalagents, schick2023toolformerlanguagemodelsteach, zhou2024webarenarealisticwebenvironment}. This misalignment makes current interfaces fragile for agent-mediated interaction and limits the reliability, verifiability, and safety of web-based agent tasks.

\section{Research Problem and Approach}
\subsection{Research Problem}
As established in Section 2.3, existing web design paradigms are structured around human-centric objectives and therefore cannot provide the integrated architectural support required to operate autonomous AI agents. A website may satisfy all conventional design criteria yet still fail to provide an appropriate and transparent environment for AI agents to complete tasks. Previous studies have given limited attention to developing a coherent framework for determining whether a website is adapted to work with an AI agent. To address this gap, this research presents a conceptual framework for agent-ready websites.
\subsection{Research Aim and Research Questions}
This study examines whether an agent-ready website design improves the effectiveness of AI browser agents at executing tasks on a website. The study compares two versions of a website: a baseline human-centered variant and an agent-ready version. The primary research questions are:
\begin{itemize}
    \item \textbf{RQ1.} What website characteristics indicate readiness for effective AI-agent interaction?
    \item \textbf{RQ2.} How does an agent-ready website affect AI browser-agent performance compared with a baseline website?
    \item \textbf{RQ3.} Across which categories of web tasks does agent-ready design yield the greatest performance improvements?
\end{itemize}

\section{Proposed Conceptual Framework: Agent-Ready Website}
Building on the failure modes identified above, this section links recurrent web-agent failures to the core dimensions of an agent-ready website. Across web-agent benchmarks and real-world studies, failures commonly manifest as observation, grounding, execution, and decision failures.
\subsection{Agent Interpretability}
Agent interpretability refers to the degree to which an AI agent can accurately parse and comprehend a website’s content, encompassing not only visible text but also page structure, HTML and Document Object Model (DOM) composition, element roles, and the organized representation of information. Within this framework, interpretability is supported by two operational sub-components: machine readability and semantic clarity.\\

Machine Readability: Machine readability pertains to the degree to which a website encodes its content and structure in formats that machines can interpret and utilize, rather than exclusively in formats designed for human visual perception. Schema markup and Schema.org vocabularies provide structured descriptions of page content and entities, helping agents identify the entities, attributes, and relationships represented on a page \cite{guha2016schema, iliadis2025schemadotorg}. Google's structured data guidelines similarly emphasize that structured data helps search systems understand and categorize page content. In practice, machine readability can be improved through the use of JavaScript Object Notation for Linked Data (JSON-LD), metadata, sitemaps, semantic HTML, and structured data concerning products or services \cite{NAVARRETE2014MET, xin2018cross, google_intro_structured_data, google_structured_data_gallery}. For automated agents, these mechanisms help reduce ambiguity about a website’s offerings and the locations of essential information. In this context, schema markup, JSON-LD, metadata, and semantic HTML should be treated not only as SEO mechanisms but also as signals that improve machine interpretability.

Semantic Clarity: Semantic clarity is how well a website conveys information, particularly in the context of agent-driven interactions. Key pages should clearly state their purpose and describe the services, products, or actions they support, as agents depend heavily on web text, HTML, and interactive components, which may be dispersed, thereby diminishing accuracy \cite{johnson2025manipulatingllmwebagents, koh2024visualwebarenaevaluatingmultimodalagents, lu2024weblinxrealworldwebsitenavigation, yang2025agentoccamsimplestrongbaseline}. Employing clear headings and comprehensive descriptions facilitates agents in accurately extracting information. Recent advancements in AI research demonstrate that HTML, DOM, accessibility features, page text, and interactive elements collectively enable agents to comprehend and engage effectively. These factors are critical in determining whether an agent can interpret content reliably.

\subsection{Agent Executability}
Agent executability refers to the degree to which an agent can reliably complete its intended actions after understanding the page, such as selecting a product, adding it to a cart, filling out a form, or navigating to the next step. Agent Actionability assesses whether a website provides transparent and reliable pathways to complete tasks. Even if the content and structure are comprehensible, tasks may prove challenging if the pathways are unclear, intricate, or misaligned with the objectives \cite{song2025browsingapibasedwebagents}. Websites should feature identifiable buttons, links, and forms with explicit relationships, enabling agents to follow procedures without confusion. Clear button labels, form fields, headings, and step-by-step workflows make these pathways easier for agents to follow. Additionally, websites can support agents through APIs or data services that provide access to up-to-date information, such as MCP, which uses standardized communication pathways \cite{hou2026mcp, ssccc}. Ultimately, Agent Executability concerns how effectively a website enables AI agents to perform tasks in a manner that is simple, transparent, and dependable.

\subsection{Agent Decision Reliability}
\begin{figure}[htbp]
  \centering
  \includegraphics[width=0.85\textwidth]{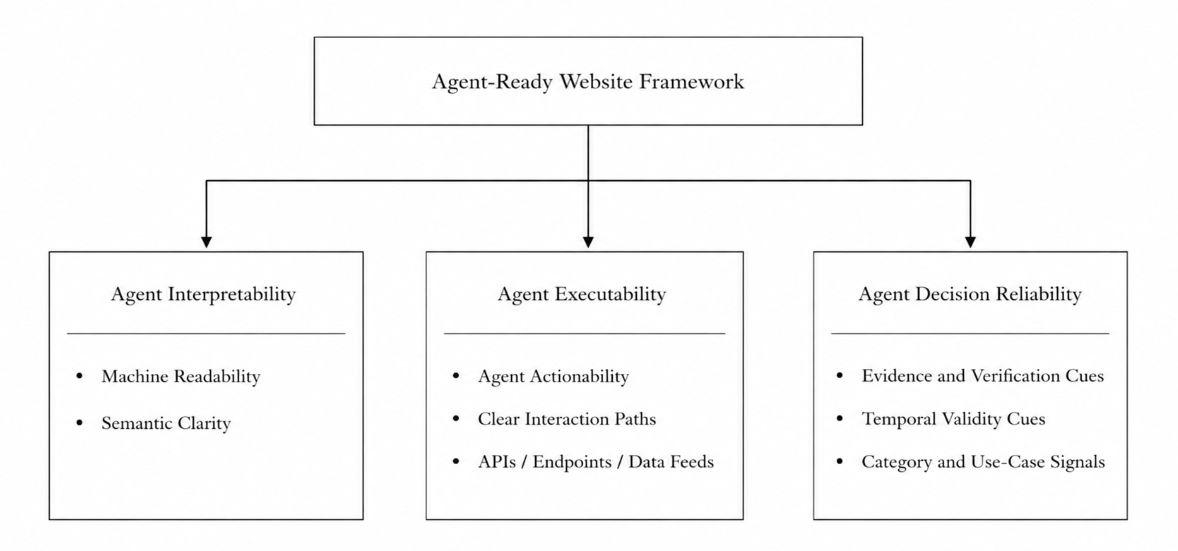}
  \caption{Conceptual structure of the agent-ready website framework}
  \label{fig:framework}
\end{figure}

Agent decision reliability refers to the degree to which a website provides signals that support dependable, up-to-date, and user-aligned decisions. Even if a site is interpretable and offers clear pathways, missing valid information can lead an agent to make recommendations, compare options, or act on incomplete or outdated data. Reliable decision-making depends on signals that help agents assess the credibility, relevance, and appropriateness of available options \cite{gautam2025multiagentsystemsmisinformationlifecycle, levy2026stwebagentbenchbenchmarkevaluatingsafety, souza2025provagentunifiedprovenancetracking, wu2025automated}. A key factor in this dimension is the availability of evidence and information that facilitate the verification and validation of claims made on the website. References, user reviews, case studies, third-party certifications, details about the organization providing the information, and similar documentation can help agents distinguish advertising claims from verifiable data. Furthermore, information must be temporally valid, as a significant portion of the data agents use on websites is subject to change over time \cite{articlessss, dai2011freshness, del2025bridging, hagele2025aspect, oche2025systematicreviewkeyretrievalaugmented}. Updating the timestamps of information on websites and ensuring its temporal validity can mitigate the risk of decisions being based on outdated information. The three dimensions agent interpretability, agent executability, and agent decision reliability are interconnected, with deficiencies in any affecting performance \cite{levy2026stwebagentbenchbenchmarkevaluatingsafety, ssccc, xue2025illusionprogressassessingcurrent}. For example, a website might present information clearly, but unclear interaction pathways can hinder task completion. Similarly, even when actions are straightforward to execute, outdated or incomplete evidence can still lead to incorrect decisions. A website's capacity to interact with AI agents is evaluated across these three dimensions, as illustrated in Figure~\ref{fig:framework}, which presents the proposed conceptual framework.

\section{Methodology}
To evaluate the practical application of the proposed framework, this study undertakes a controlled proof-of-concept experiment comparing two versions of an e-commerce website. The design, catalog, pricing, inventory conditions, tasks, and runtime conditions were held constant, with the only variation being the integration of additional features outlined in the framework into the agent-ready version.

\subsection{Experimental Task Set and Website Variants}
This experiment used two versions of an e-commerce platform: Website A as the baseline and Website B as the agent-ready variant. Both shared the same structure, products, pages, pricing, inventory, ratings, and checkout procedures, differing only in Website B's agent readiness criteria, as specified in Table~\ref{tab:website-variants}. Five web tasks were conducted to evaluate components of the framework, including actions, information retrieval, and decision-making. Task 5 was designed as a lower-complexity policy-retrieval task to test whether agent-ready features also affect routine information-access scenarios. All tasks used the same prompts and parameters, and results are available on GitHub to ensure reproducibility.

\begin{table}[htbp]
  \centering
  \caption{Comparison of the baseline and agent-ready website variants}
  \label{tab:website-variants}
  \footnotesize
  \begin{tabular}{@{}p{3.5cm}p{5.2cm}p{5.2cm}@{}}
    \toprule
    \textbf{Design Aspect} & \textbf{Baseline (Website A)} & \textbf{Agent-Ready (Website B)} \\
    \midrule
    Product data exposure &
      Product data embedded in JavaScript &
      Product data also exposed via JSON files and JSON-LD \\
    \addlinespace
    Semantic labels &
      Some controls include basic labels or attributes &
      More elements include \texttt{aria-label}, \texttt{data-*} attributes, and explicit identifiers \\
    \addlinespace
    Action controls &
      Cart and product actions rely on visible button text and handlers &
      Actions include explicit \texttt{data-*} attributes, product IDs, names, stock, and availability fields \\
    \addlinespace
    Decision reliability signals &
      Reviews, stock text, and basic policy information visible on page &
      Evidence files/pages, temporal fields, and clearer product/use-case information added \\
    \bottomrule
  \end{tabular}
\end{table}

\subsection{Experiment Setup and Run Protocol}
The experiments were conducted using the open-source agentic web framework browser-use\footnote{\url{https://github.com/browser-use/browser-use}}. Three models, GPT-4.1, Gemini 2.5 Flash, and Grok 4 Fast, were evaluated with temperature set to 0 and top p set to 1. Each session was limited to thirty steps and up to three errors. To ensure independence between runs, the local server was restarted before each trial, and a temporary browser profile was created at the outset and deleted upon completion. Each of the five tasks was performed ten times per model across both website versions, resulting in three hundred independent runs. Interaction history, token consumption (where available), errors, and execution outputs were recorded in the test logs for each run.

\subsection{Evaluation Scoring and Analysis}
Agent performance was evaluated by two annotators who reviewed the agent's outputs, including terminal logs and the model's JSON output, and classified each run as PASS, PARTIAL, or FAIL. A PASS label was assigned when all primary task requirements were fully met; this was counted as both a Strict Success and a Functional Success. A PARTIAL label was assigned when the main task objective was achieved but a non-critical or moderate issue remained; this was considered a Functional Success but not a Strict Success. A FAIL label was assigned when the main task objective was not achieved, or an important instruction was violated. In addition to the outcome, a separate error field was recorded for runs that contained errors. Step count and token consumption were also recorded. Results were then aggregated and reviewed at the per-task level.

\section{Results and Analysis}\label{sec:results}
\subsection{Overall Task Success}
Table~\ref{tab:model-task-outcomes} demonstrates that the agent-ready website surpassed the baseline across all models, with 134 successful passes out of 150 (89.3\%) compared to 74 (49.3\%). The difference amounted to 40 percentage points. A chi-square test confirmed a significant association between website type and outcome (\(\chi^{2}(2, N=300)=60.79, p<0.001\)), with Cramer's V at 0.45 indicating a moderate effect. The agent-ready version generally outperformed the baseline, although improvements varied. Nevertheless, the baseline condition still produced functional outcomes in several runs. However, agents often achieved only a broadly correct answer while failing to extract complete information, provide sufficient evidence, or fully adhere to constraints. This was common in information extraction, product comparison, and multi-constraint tasks. The agent-ready website did not eliminate all errors; these were primarily observed in complex tasks and resulted from incorrect product choices, overlooked constraints, or incorrect conclusions. Overall, the agent-ready system enhances task reliability and quality and reduces partial outcomes, but it cannot entirely rectify reasoning errors or model limitations.

\begin{table}[htbp]
  \centering
  \caption{Model-level task outcomes across website variants}
  \label{tab:model-task-outcomes}
  \footnotesize
  \begin{tabular}{@{}llrrrrr@{}}
    \toprule
    \textbf{Model} & \textbf{Website} & \textbf{PASS} & \textbf{PARTIAL} & \textbf{FAIL} & \textbf{Strict success} & \textbf{Functional success} \\
    \midrule
    GPT-4.1 & Agent-ready & 43 & 0 & 7 & 86\% & 86\% \\
    GPT-4.1 & Baseline & 16 & 18 & 16 & 32\% & 68\% \\
    Gemini-2.5 Flash & Agent-ready & 44 & 2 & 4 & 88\% & 92\% \\
    Gemini-2.5 Flash & Baseline & 26 & 11 & 13 & 52\% & 74\% \\
    Grok-4 Fast & Agent-ready & 47 & 1 & 2 & 94\% & 96\% \\
    Grok-4 Fast & Baseline & 32 & 14 & 4 & 64\% & 92\% \\
    \midrule
    Total & Agent-ready & 134 & 3 & 13 & 89.3\% & 91.3\% \\
    Total & Baseline & 74 & 43 & 33 & 49.3\% & 78\% \\
    \bottomrule
  \end{tabular}
\end{table}

\subsection{Task Results and Main Failure Patterns}
Table~\ref{tab:task-pass-rates} presents the differences in PASS rates across all three models for each task. The results show that the agent-ready website increased the success rate on all five tasks. However, the improvement was smaller for Task~5, as the baseline version also performed relatively well on this task. The greatest improvements were observed in Tasks~1, 2, and 3, which require precise information extraction, comparison, and selection under multiple constraints. Model-level PASS, PARTIAL, and FAIL outcomes are provided in Table~\ref{tab:model-task-outcomes}, while task-level PASS rate comparisons and Fisher's exact test results are summarized in Table~\ref{tab:task-pass-rates}. Qualitative error patterns are discussed below. Task~4 was challenging for both versions because it required satisfying multiple constraints simultaneously; nevertheless, the agent-ready version increased the PASS rate by 16.7 percentage points. Task~5 showed a smaller difference, consistent with its simpler, policy-based nature. To assess per-task statistical significance, we conducted Fisher's exact tests comparing PASS versus non-PASS outcomes (PARTIAL and FAIL combined) for each task. Results showed statistically significant improvements for Tasks~1, 2, and 3 (all \(p<0.05\)), whereas Tasks~4 and 5 did not reach significance (Table~\ref{tab:task-pass-rates}).

\begin{table}[htbp]
  \centering
  \caption{Compact task-level PASS rate comparison}
  \label{tab:task-pass-rates}
  \footnotesize
  \begin{tabular}{@{}lcccccc@{}}
    \toprule
    \textbf{Task} & \textbf{} & \textbf{Baseline PASS rate} & \textbf{Agent-ready PASS rate} & \textbf{Difference} & \textbf{P} \\
    \midrule
    Task~1 & & 73.3\% & 96.7\% & +23.4 pp & 0.026 \\
    Task~2 & & 23.3\% & 100\% & +76.7 pp & <0.001 \\
    Task~3 & & 16.7\% & 93.3\% & +76.6 pp & <0.001 \\
    Task~4 & & 60.0\% & 76.7\% & +16.7 pp & 0.267 \\
    Task~5 & & 73.3\% & 80.0\% & +6.7 pp & 0.761 \\
    \bottomrule
  \end{tabular}
\end{table}

\subsection{Error Pattern Analysis}
Table~\ref{tab:model-task-outcomes} summarizes model-level task outcomes across website variants. Detailed task-level error patterns are discussed qualitatively below. The baseline exhibited common errors, including incomplete data extraction (Task~2), partial comparisons (Task~3), incorrect product selections, unmet constraints (Task~4), and incomplete interpretations of store policies (Task~5). Errors in Task~1 were more model-dependent, involving inventory assessment errors, incorrect product selections, or shopping cart issues. The agent-ready version reduced both error rates and diversity, particularly in complex tasks, indicating its effectiveness in reducing access and extraction errors. Some failures in reasoning and integration persisted. The reduction in PARTIAL outcomes was significant: in the baseline condition, agents frequently identified the correct page or product but failed to extract complete information or draw fully supported conclusions. Conversely, the agent-ready version exhibited fewer partial outcomes, attributable to clearer fields, explicit labels, structured data, and unambiguous information, which helped the agent better fulfill the task requirements.

\subsection{Step Count and Token Usage}
Average step count served as a secondary indicator of execution effort, not a primary success criterion. The average decreased from 9.31 steps in the baseline condition to 6.49 in the agent-ready condition, a 30.4\% reduction. The reduction was observed across all three models, with the largest absolute decrease for Gemini 2.5 Flash; model-level details are reported in Table~\ref{tab:model-step-token}. The largest reductions occurred in Tasks~3 and 4, while Task~5 showed almost no change, supporting the interpretation that the agent-ready design primarily reduced interaction effort in extraction and comparison. Prompt token consumption followed a consistent pattern, decreasing across all three models: by 18.72\% for Gemini 2.5 Flash, 40.47\% for Grok 4 Fast, and 37.43\% for GPT-4.1. Full model-level token totals are also provided in Table~\ref{tab:model-step-token}.

\begin{table}[htbp]
  \centering
  \caption{Model-level step count and token consumption across website variants}
  \label{tab:model-step-token}
  \footnotesize
  \begin{tabular}{@{}lrrrr@{}}
    \toprule
    \textbf{Model} & \textbf{Avg. Steps Agent-ready} & \textbf{Avg. Steps Baseline} & \textbf{Token Usage Agent-ready} & \textbf{Token Usage Baseline} \\
    \midrule
    GPT-4.1 & 3.70 & 5.96 & 1,770,532 & 2,829,720 \\
    Gemini-2.5 Flash & 11.56 & 15.02 & 4,958,307 & 6,099,907 \\
    Grok-4 Fast & 4.20 & 6.96 & 2,189,279 & 3,677,702 \\
    \bottomrule
  \end{tabular}
\end{table}

\section{Discussion}
\subsection{Response to Research Question 1: Defining agent readiness}
A website is better equipped to interact with AI agents when it presents information, action pathways, evidence, and relevant conditions in a way that allows agents to identify, interpret, validate, and utilize them easily. As detailed in Section~4, agent-readiness is defined across three dimensions: agent interpretability, agent executability, and agent decision reliability. While these dimensions do not cover every aspect of agent-readiness, the evidence supports the view that clearer website-level signals reduce ambiguity for browser agents, particularly in tasks that require complete information extraction, option comparison, or selection under multiple constraints.

\subsection{Response to Research Question 2: Effect on AI agent performance}
Consistent with the experimental results reported above, this study demonstrated that, under the controlled conditions of this experiment, implementing the agent-ready framework improved browser agent performance across all three models. The improvement was consistent across all three models. As shown in Table~\ref{tab:model-task-outcomes}, the agent-ready design not only increased the strict success rate but also substantially improved the quality and completeness of previously incomplete outputs. A similar pattern was observed in efficiency measures (see Table~\ref{tab:model-step-token}): agents required fewer redundant actions, less error-recovery effort, and fewer repeated attempts at information extraction.

\subsection{Response to Research Question 3: Variation across tasks and models}
The effect of agent-ready design was most pronounced in more complex, multi-step tasks. In tasks requiring repeated navigation, option comparison, simultaneous satisfaction of multiple constraints, or execution of operations on the website, the agent-ready version consistently produced more complete outcomes. This suggests that clearer structure, task-relevant fields, and more explicit action cues reduced ambiguity and helped agents follow more complex paths with fewer errors. Agent-ready design did not, however, eliminate all errors. Some failures occurred even when the required information was available, and the page structure was clearly presented, indicating that these errors were attributable to the agent's own reasoning process, selection strategy, or final interpretation rather than to missing information or poor website structure. This highlights that while agent-ready design can support agents, it cannot compensate for all errors arising from model-level limitations. Task~5 further supports this finding: adding agent-ready capabilities did not introduce additional complexity for agents, and its primary value was apparent in tasks requiring structured information extraction, option comparison, or multi-step execution.

\section{Conclusion}
The findings suggest that agent readiness should be treated as a website-level design capability, distinct from but complementary to human usability, SEO, structured data, and accessibility. To operationalize this capability, the paper introduces a conceptual framework structured around three dimensions: agent interpretability, agent executability, and agent decision reliability. A controlled proof-of-concept experiment provides preliminary evidence of the practical significance of these design choices: the agent-ready website achieved an 89.3\% success rate, compared with 49.3\% for the baseline, with the most notable improvements observed in tasks involving structured extraction, option comparison, and multi-constraint selection. Furthermore, the agent-ready version reduced the number of PARTIAL outcomes from 43 to 3. It decreased the average number of steps by 30.4\%, indicating that clearer structural organization and explicit action cues mitigate both incompleteness and interaction effort. It is important to note that the framework does not guarantee correct agent behavior, as some errors persisted, particularly in tasks requiring multi-step reasoning rather than mere information access. Its primary contribution lies in identifying website-level characteristics that enhance the reliability and efficiency of agent interactions. For e-commerce managers and platform designers, these findings imply that as AI-mediated traffic expands, agent readiness should be considered alongside human usability and SEO as a fundamental aspect of design and quality management. Future research should aim to extend this framework to real-world websites, encompass broader domains, and incorporate a wider spectrum of agent systems.

\section{Limitations}
This study is a controlled proof of concept rather than a full validation of the proposed framework, and the results should not be generalized to all domains, websites, or agent systems. Future work should evaluate the framework on real-world websites across broader domains, with a wider range of agent systems, and through ablation studies that isolate the contribution of individual agent-ready features.

\section*{Acknowledgments}
This research received no specific grant from any funding agency in the public, commercial, or not-for-profit sectors.

\bibliographystyle{unsrtnat}
\bibliography{references}

\end{document}